\pdfoutput=1
\documentclass[letterpaper, 10 pt, conference]{ieeeconf}
\IEEEoverridecommandlockouts                              

\usepackage{graphics} 
\usepackage{epsfig} 
\usepackage{amssymb}
\usepackage{xcolor}
\definecolor{peach}{RGB}{255,220,102}

\usepackage[percent]{overpic}

\newcommand*\circled[1]{\tikz[baseline=(char.base)]{
            \node[shape=circle,draw,inner sep=1pt] (char) {#1};}}

\usepackage{transparent}

\usepackage{booktabs}
\newcommand{\ra}[1]{\renewcommand{\arraystretch}{#1}}

\usepackage{caption, copyrightbox}

\newcommand{\Hquad}{\hspace{0.1cm}} 

\title{\LARGE \bf
Full-Scale Continuous Synthetic Sonar Data Generation with Markov Conditional Generative Adversarial Networks*
}

\author{Marija Jegorova$^{1}$, Antti Ilari Karjalainen$^{2}$, Jose Vazquez$^{2}$, Timothy Hospedales$^{1}$
\thanks{* This work was supported by SeeByte Ltd}
\thanks{$^{1}$ University of Edinburgh, UK
        {\tt\small m.jegorova@ed.ac.uk, t.hospedales@ed.ac.uk}}%
\thanks{$^{2}$ SeeByte Ltd., UK
        {\tt\small antti.karjalainen@seebyte.com, jose.vazquez@seebyte.com}}%
}

\begin{document}

\maketitle{}
\thispagestyle{empty}
\pagestyle{empty}

\begin{abstract}

Deployment and operation of autonomous underwater vehicles is expensive and time-consuming. High-quality realistic sonar data simulation could be of benefit to multiple applications, including training of human operators for post-mission analysis, as well as tuning and validation of autonomous target recognition (ATR) systems for underwater vehicles. Producing realistic synthetic sonar imagery is a challenging problem as the model has to account for specific artefacts of real acoustic sensors, vehicle attitude, and a variety of environmental factors. We propose a novel method for generating realistic-looking sonar side-scans of full-length missions, called Markov Conditional pix2pix (MC-pix2pix). Quantitative assessment results confirm that the quality of the produced data is almost indistinguishable from real. Furthermore, we show that bootstrapping ATR systems with MC-pix2pix data can improve the performance. Synthetic data is generated 18 times faster than real acquisition speed, with full user control over the topography of the generated data.

\end{abstract}
\setlength{\belowcaptionskip}{-10pt}
\section{INTRODUCTION}
In underwater environments, sonars are often preferred over other sensors due to the high density of organic material and inorganic dust that can restrain optical visibility.
Because of their perceptual robustness, sonar sensor data is heavily relied upon for tasks such as object localization, oil-pipe and infrastructure inspections, search and rescue, and other commercial and military applications.

A vast amount of data is required to construct detection and recognition models for automating most of these applications. Underwater data collection is expensive, time-consuming, and in most cases commercially sensitive. A means of synthetically creating such data would be highly beneficial to the underwater sensor processing community, as it would mitigate the costly process of data collection by instead making better use of the available real training data.

Existing techniques for image synthesis, such as generative adversarial networks (GANs)~\cite{vanillaGAN} have recently grown capable of producing and enhancing images of high resolution (e.g. 2048$\times$1024 by pix2pixHD \cite{pix2pixHD}). However, typical underwater survey missions sonar images usually exceed the image resolution of 300,000$\times$512 pixels. 
We propose the Markov Conditional pix2pix (MC-pix2pix) method which, to our knowledge, is the first method capable of generating realistic sensory output for full-length missions, given a small amount of initial training data. Crucially, such generation runs 18 times faster than acquisition on the real hardware, resulting in a realistic and faster than real-time simulator.

To demonstrate the utility of our approach, we provide quantitative results in two extrinsic evaluation tasks: (i) the synthetic data is almost impossible to distinguish from real data for domain experts, thus enabling training of teleoperators without using real hardware; (ii) significant performance gains are achieved when using this synthetic data to augment training datasets for autonomous target recognition (ATR) in a variety of seabed conditions. The results presented in this work are produced with Marine Sonic sonar side-scan data, but the method itself is sonar-agnostic.


   \begin{figure*}[t]
      \centering
      \vspace{0.2cm}
      \begin{overpic}[scale = 0.174]{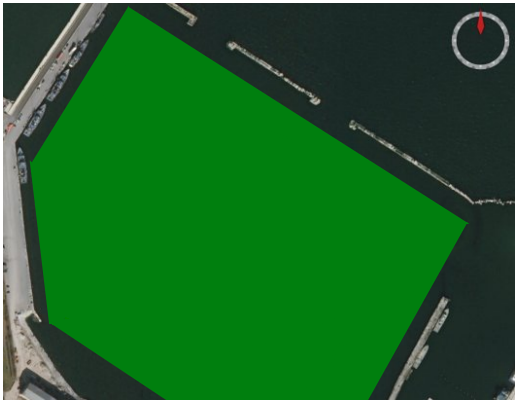}\put(3,66){\textbf{\textcolor{white}{\circled{1}}}}\end{overpic}\begin{overpic}[scale = 0.174]{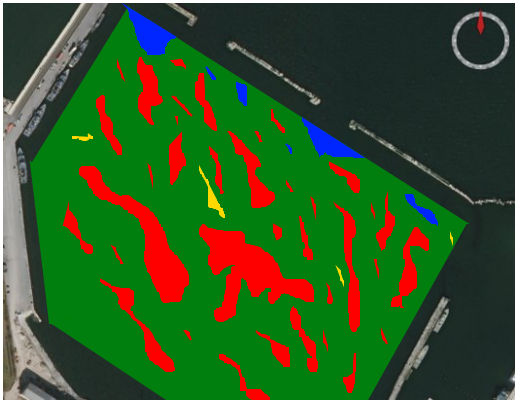}\put(3,66){\textbf{\textcolor{white}{\circled{2}}}}\end{overpic}\begin{overpic}[scale = 0.174]{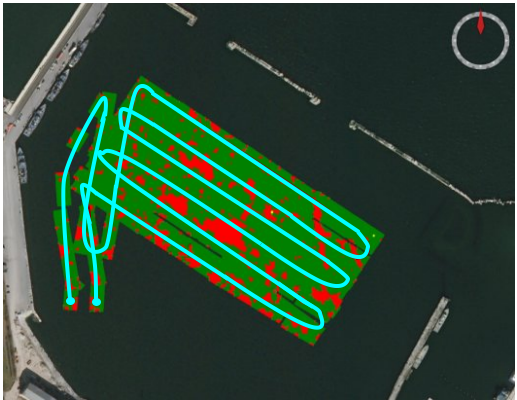}\put(3,66){\textbf{\textcolor{white}{\circled{3}}}}\end{overpic}\begin{overpic}[scale = 0.0905]{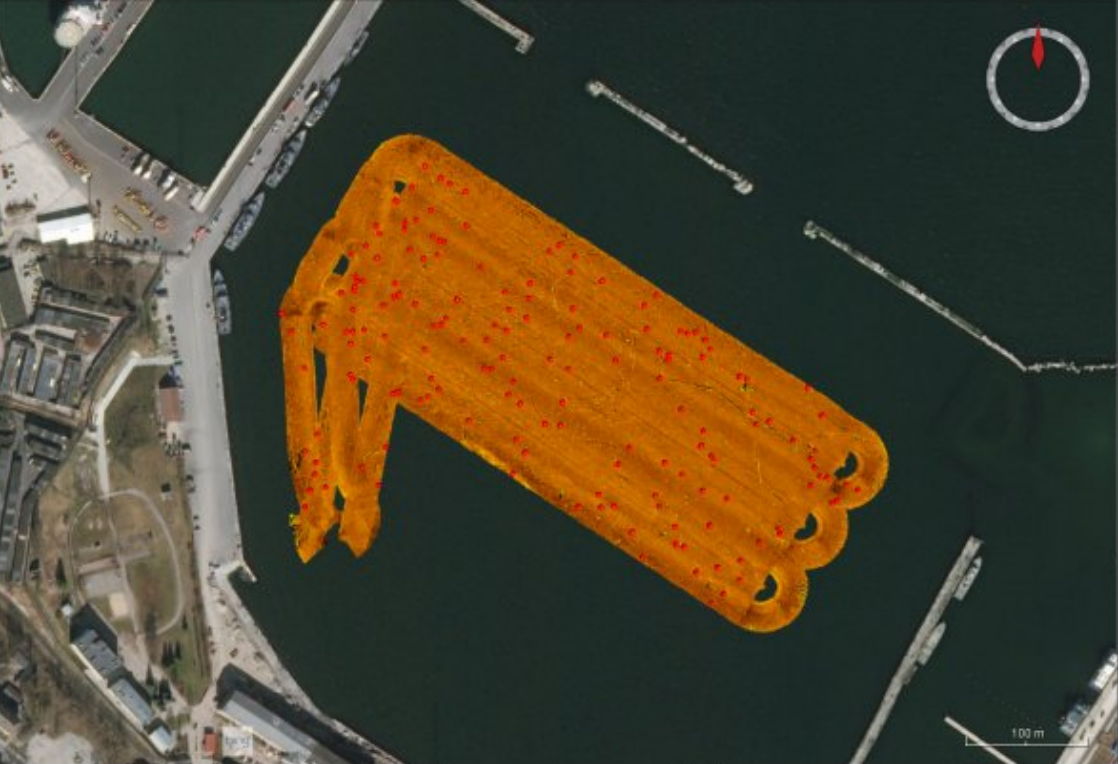}\put(3,60){\textbf{\textcolor{white}{\circled{4}}}}\end{overpic}\begin{overpic}[scale = 0.083]{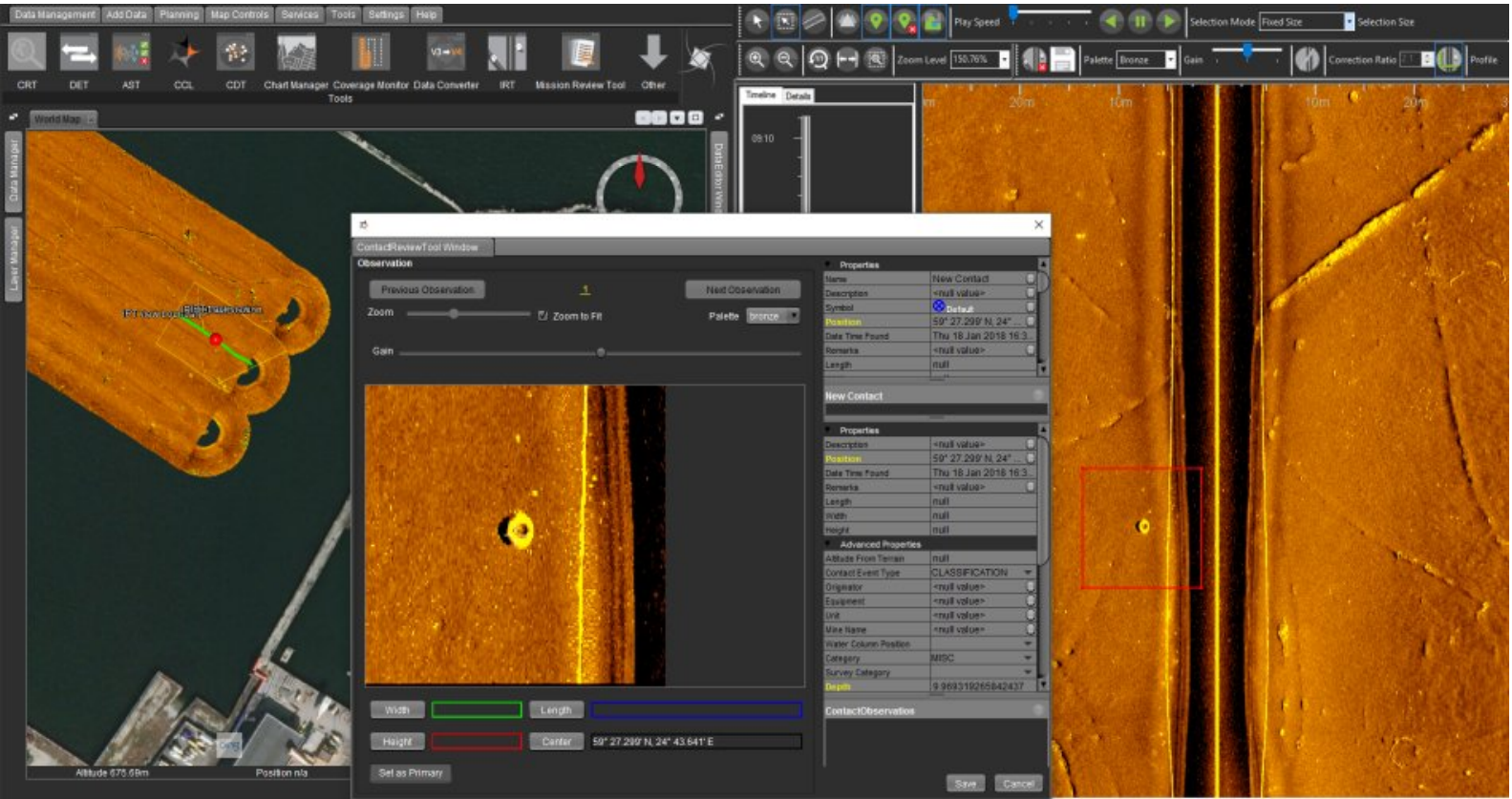}\put(3,46){\textbf{\textcolor{white}{\circled{5}}}}\end{overpic}
      \caption{\textbf{Pipeline:} 1-2. Training instructor labels the map regions with desired textures and target locations. 3. Trainee operator creates a route across a given map with an objective of collecting data for locating hidden targets. 4. The model receives semantic maps and route as inputs and outputs synthetic sonar data for the entire mission. (In real life vehicle completes the mission delivering the sonar data collected.) 5. Example of sonar images when inspected by a human operator.}
      \label{pipeline}
   \end{figure*}

\vspace{-3.5 pt}

\section{RELATED WORK}

GANs~\cite{vanillaGAN} are a class of neural network models for the realistic data generation. Since their initial introduction in 2014, a large number of extensions have been proposed for various applications, primarily focused on realistic image and video generation \cite{LargeScaleGAN, SuperGAN, text2image, videoGAN, vid2vid}, where only a limited amount of training data is available. In contrast to these tasks, there is comparatively little work investigating how GANs can be of benefit in robotics. Although robots that use image recognition in domains where training data is scarce may benefit from the conventional applications of GANs. Some applications that are more relevant to robotics include GAN-based approaches to imitation learning~\cite{GAIL, infoGAIL}, which allow robots to efficiently learn a single policy or a discrete set of policies from demonstration, and direct generation of robot control policy repertoires~\cite{GPN}, a technique that enables sampling from the continuous target-conditional distributions over the control policies within a scope of a given task.

Facilitation of the user-controlled simulation requires some form of the information transfer. GANs have been extensively used for style transfer and image-to-image translation, beginning with cycleGANs for transfer between unpaired images \cite{CycleGAN2017}. However, on paired image translation problems -- the task we are primarily concerned with -- pix2pix \cite{pix2pix2016} and its subsequent variations \cite{pix2pixHD, SPADE2019} are known to perform considerably better. No current image translation methods can be directly applied to full-mission sonar data because of the extremely high resolution. The size of the full image to be generated is usually in excess of 300,000$\times$512 pixels -- roughly the amount of data generated by a short two hour training mission. Our method solves this problem by producing such image in a piece-wise sequential manner, and ensures continuity of the output through the use of a Markov assumption. The use of Markov assumption here is justified by the temporal nature of the real data acquisition during real mission, as well as by the general spatio-temporal continuity of the required data.

The previous attempt of generating sequential data with GANs, recurrent GANs (RGANs) and recurrent conditional GANs (RCGANs)~\cite{RGAN_RCGAN}, focused on medical sequence generation. This has been accomplished through the use of recurrent neural network architectures. There are two issues with applying these techniques to the problem of synthetic sonar imagery generation. Firstly, we require the control over the topography. So the model architecture would need to be modified for image translation with convolutional layers, which would further require one to use backpropagation through time for training the network, rendering the training process computationally intractable for the size of data that we work with. Secondly, RGAN and RCGAN are designed to produce semantically realistic sequences, whereas we require perceptually realistic image sequences. 
Additionally, the nature of the sonar imagery suggests that the Markov assumption alone is enough for the coherency and continuity.

A small number of papers address the underwater robot perception problems with GANs: the work of~\cite{SSPDcycleGAN} shows cycleGANs enhancing synthetic target objects for embedding them into the real sonar images in order to train an ATR system, while \cite{UGAN} proposes a method for refining video images rather than generating new acoustic imagery.

Until now the applications of GANs to underwater sensory data were mostly enhancing the imagery, either optical or acoustic, rather than generating brand new data. MC-pix2pix is also the first model of its type addressing the generation of a whole mission's worth of data rather than smaller images.


\section{PROBLEM AND MOTIVATION}
\subsection{Why generate sonar images?}
The key application for high-quality simulation is bootstrapping autonomous target detection and recognition (ATR) methods when training data is scarce, or some types of seabeds are underrepresented in the real training set. In section~\ref{atr_section} we show improvements in the ATR performance when generating a variety of seabeds
and introducing them into the ATR training together with the available real data.

Realistic simulation could also benefit the training of teleoperators for mission planning and interpretation of sonar imagery, replacing the costly real data collection.

\subsection{Synthetic framework for training of the vehicle operators}

The simulation pipeline, presented in the Figure~\ref{pipeline}, assumes that the training instructor marks the regions of the map with a specific topography, such as rocks, ripples, clutter, and objects of interest. The trainee operator is presented with this map without the target objects marked, and creates a route over it that should allow the robot to locate these hidden objects. Given semantic maps provided by an instructor and the route created by trainee operator, the purpose of our technique is to generate realistic seabed scans for the entire mission. Methods such as \cite{SSPDcycleGAN} can be employed to embed the target objects in the requested locations in the synthetic seabed scans, and can then be displayed to the trainee operator for visual inspection and object detection, just like during a real post-mission analysis.

The emphasis of this work is on keeping synthetic data maximally consistent and realistic, whilst achieving the highest generation speed possible.

\subsection{Problem Specification}
For the task described above the following requirements should be considered: 

\begin{itemize}
    \item Realistic looking synthetic data generation: the main focus of this work. Our method is based on GANs because they have been identified as the current best approach for generating realistic imagery.
    \item Spatial coherency: imagery of the entire mission should appear continuous and consistent. The paired nature of pix2pix guarantees consistency within topographical features represented as different labels in semantic maps. Additional conditions are introduced in section~\ref{method}, and improve the continuity of the output further.
    \item Viewpoints invariance: the same section of the map should appear texture-consistent when observed from different viewpoints.
    \item Speed of generation: in practice, for faster than real-time simulations, our model should be significantly faster at test time than real-time sonar data acquisition.
\end{itemize}

   \begin{figure*}
      \centering
      
      
    \begin{overpic}[scale = 1]{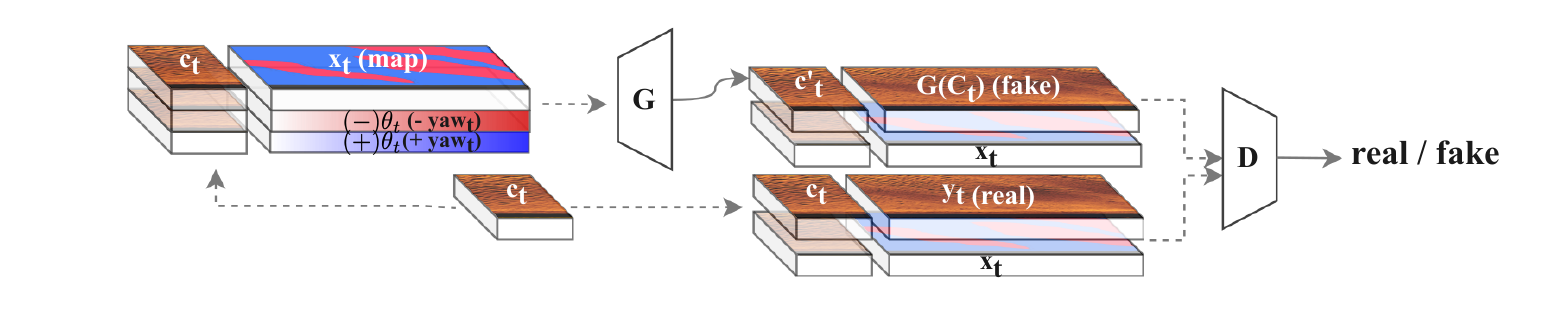}\put(-6.5, 16.5){\textbf{\textcolor{gray}{Training time:}}}\end{overpic}\vspace{0px}
    \begin{overpic}[scale = 1.15]{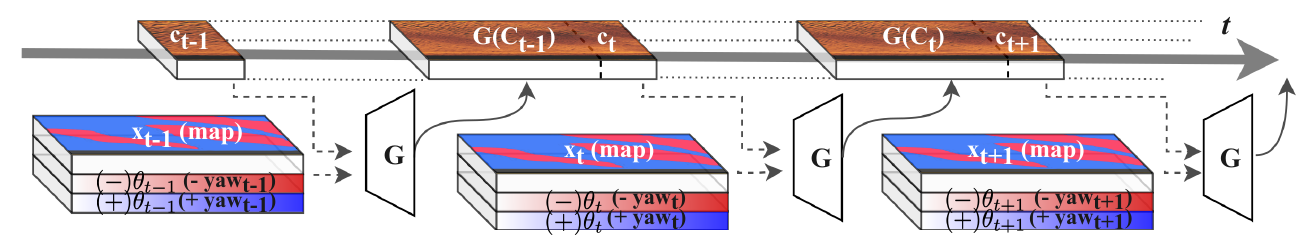}\put(-8,17){\textbf{\textcolor{gray}{Test time:}}}\end{overpic}
    
      
      \caption{\textbf{Training time:} similarly to the pix2pix, the generator inputs semantic maps corresponding to the desired topography $x_t$, outputs synthetic sonar-scan data $G_t(C_t)$. It is extended to accept two additional conditions - a snippet of the previous image $c_{t}$ facilitating continuity in the generated mission and yaw indicating the requested turns of the vehicle. Output is then labelled by discriminator as real or fake along with the real images. \textbf{Test time:} at each time-step, the generator processes a semantic map of requested configuration, yaw variable (responsible for turn distortions, defined by the vehicle trajectory), and a small snippet of the previous synthetic image to enforce the continuity of the seabed throughout the mission.}
      \label{architecture}
   \end{figure*}

\section{EXPERIMENTAL SETUP}
\subsection{Training Data}
The real side-scan sonar data used for the experiments is acquired with a Marine Sonic sonar. Its across-track resolution is 512 pixels ($\times 2$ for both port and starboard channels). The vehicle travels at the speed of 1 meter per second and generates approximately 16 pings per second. As the vehicle turns it causes distortions in the images, and models that generate synthetic imagery should be expected to produce similar distortions. Our model accounts for these distortions using the desired vehicle attitude information (yaw, pitch, and roll). Only yaw information is provided by the gathered training data, but we note that our method is able to incorporate pitch and roll data as well.


To create a training dataset, sonar scans were sliced into 464$\times$512 images. Our model was trained on a relatively small dataset of 540 of these images (and their corresponding semantic maps). Increasing the training set size might bring further improvements but in our experience this method works with as few as 200 training image samples. 

\vspace{-4px}
\subsection{Assessment metrics}
\begin{figure*}
  \flushleft
  \vspace{0.2cm}
  {\textcolor{gray}{$\Hquad\Hquad\Hquad$1. Semantic Maps$\quad\quad\Hquad\Hquad\Hquad$2. pix2pix$\quad\quad\quad\quad\quad\quad$3. pix2pix-blended$\quad\quad$4. MC-pix2pix (ours)$\quad\Hquad\Hquad$5. Real Side-Scans}}
  
  \centering
  \begin{overpic}[scale=0.195]{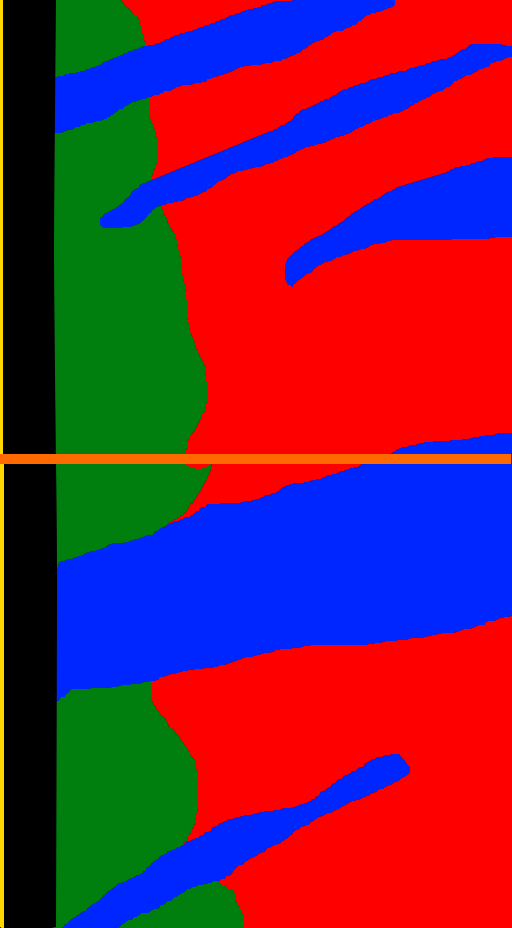}\put(8,2){{\includegraphics[scale=0.45]{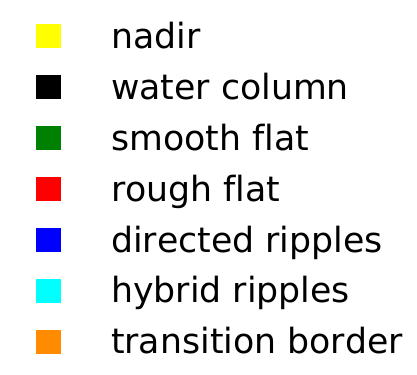}}}\end{overpic}\hspace{0.0cm}\includegraphics[scale=0.195]{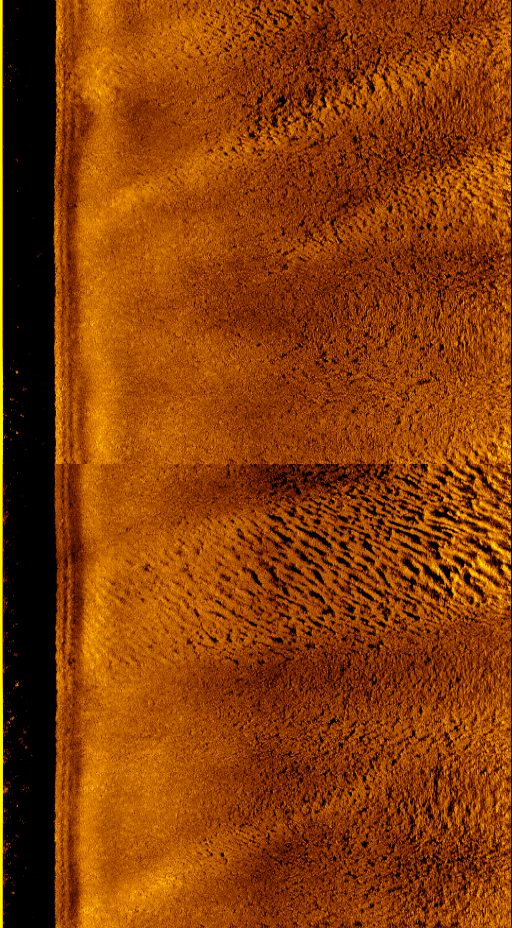}\hspace{0.0cm}\includegraphics[scale=0.195]{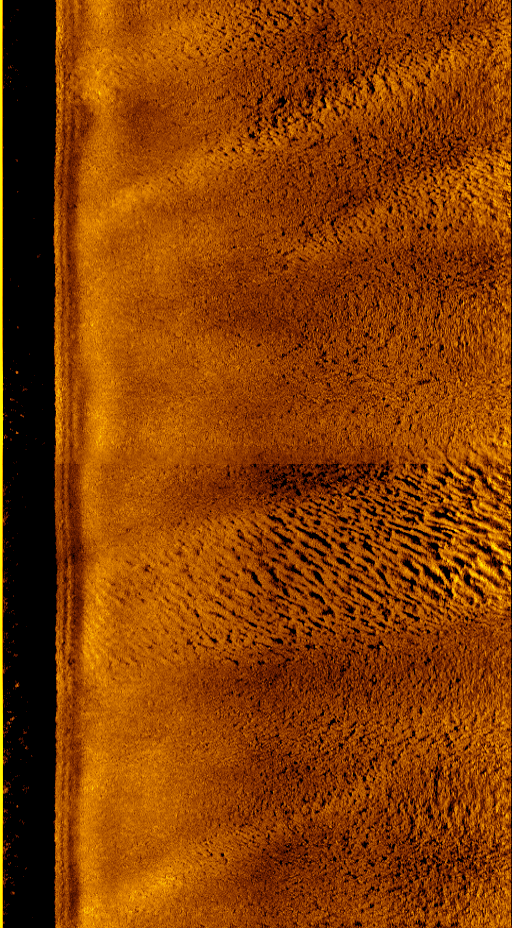}\hspace{0.0cm}\includegraphics[scale=0.195]{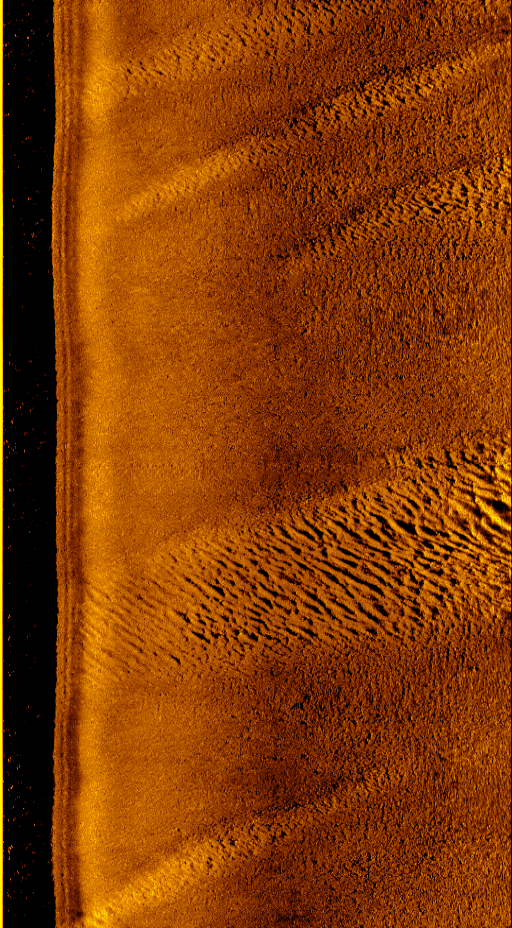}\hspace{0.0cm}\begin{overpic}[scale=0.195]{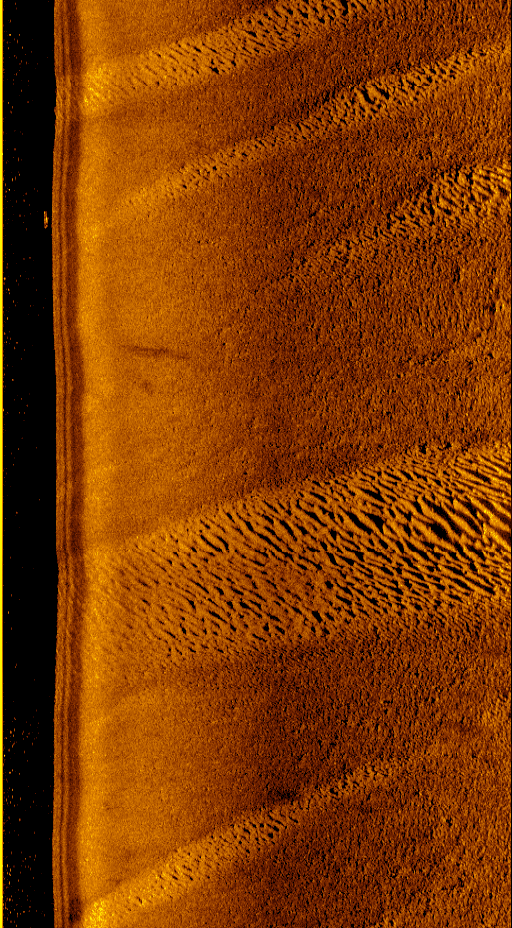}\put(32,7){\includegraphics[scale=0.09]{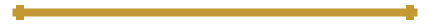}}\put(35, 3){\footnotesize{\textcolor{peach}{10 meters}}}\end{overpic}
  
  
  
  \caption{\textbf{Visual Comparison (left-to-right): 1.} Semantic maps are used as an input by all of the compared models. In reality the semantic maps are grey-scale with different shades corresponding to different types of terrain. Colour is introduced for visualisation purposes only. Border-line label indicates the transition between the images for convenience of the reader and is not present in the input of the model. \textbf{2.} Original pix2pix example has a clear sharp transition border between the images (in the middle). This is because the image patterns or intensities are not shared between adjacent images. \textbf{3.} pix2pix with sigmoid-smoothing applied at the transition demonstrates that simple post-processing is not particularly good at matching the textures of the seabeds. \textbf{4.} MC-pix2pix (ours) has clearly smoother transition border, enabling it to produce continuous imagery for missions of any length when run repeatedly. \textbf{5.} The real data example. }
  \label{results_visual}
\end{figure*}

In addition to the visual examples provided in Figure~\ref{results_visual}, the model performance has also been quantitatively assessed using the following metrics:
\begin{itemize}
    \item Human visual assessment score: we provide the statistics on distinguishing real sonar imagery from the synthetic images. It is collected from 30 participants with a variety of experience of working with underwater sonar data. During the test, participants were allowed to inspect images without the time limit.
    \item Fresch\'et Inception Distance \cite{FID2017}: often used for quality assessment of generated images, this is a heuristic for measuring the difference between the real and synthetic image distributions.
    \item Generation speed at test time: crucially for practical application, generative models must provide results of the requested quality without compromising the speed of generation. A minimal requirement is an order of magnitude faster than real data collection speed.
    \item Performance improvements of an ATR detection algorithm when bootstrapped with the synthetic data along with the available real data. This is assessed in terms of mean Average Precision (mAP) and F1-score - harmonic mean between the precision and the recall of the ATR system. \label{f1} 
\end{itemize}

\section{METHOD AND ARCHITECTURE}\label{method}
An overview of the model architecture is provided in Figure~\ref{architecture}\footnote{Please note: semantic maps, sonar-scans, and yaw variables are single-channel and are only coloured as RGB for illustration purposes.}. It resembles the fully-convolutional pix2pix architecture with 9 resnet blocks \cite{pix2pix2016}. Importantly, it is designed to accept two conditions at the input level \cite{conditionalGAN}.

\subsubsection{Conditions} the first condition, $c_{t-1}$, enforces the visual continuity of the generated output at test time. The information is conveyed by a short snippet taken from the end of the previous image, enabling the model to run self-conditionally at test time, as illustrated in Figure~\ref{architecture}.

The second condition, a yaw-based metric $\theta_t$, takes care of the image distortions caused by turns, it is calculated as: 
$$
\theta_t = 5\hspace{0.5em} max( 1, | \psi_t - \psi_{t+50} |)      \eqno{(1)}
$$
where $\psi_t$ is the yaw for ping $t$, and the sign of $(\psi_t - \psi_{t+50})$ is used to determine whether the clockwise or counterclockwise turn is expected. $\theta_t$ is calculated per ping (per row) of the corresponding semantic map $x_t$, the resulting vector gets repeated column-wise, separated into two arrays based on the sign of $(\psi_t - \psi_{t+50})$, and overlayed with the single-channel semantic map $x_t$, completing the generator input.

\subsubsection{At training time} the generator inputs the  single-channel semantic maps $x_t$ and the two conditions - yaw variable $\theta_t$ and the previous image snippet $c_t$.
The generator outputs single-channel generated sonar images $G(x_t, ...)$. the discriminator receives all available data except the yaw variable - semantic maps $x_t$, condition $c_t$, and real images $y_t$, and generated images $G(x_t, ...)$. The discriminator outputs the verdict on whether the image is real or fake. The discriminator is rewarded based on how well it can distinguish the synthetic image $G(x_t, ...)$ from real $y_t$, generator - based on if it managed to "fool" the discriminator.

This model is adversarially trained for 200 epochs with batch-size 10 and 3 repetitions of discriminator for 1 of generator per each epoch with the following loss function:

\[  {G_t}^\star = \arg \min_G \max_D \big\{ \mathbb{E}_{x_t, y_t}[log D(x_t, y_t)] \]
\[\quad\quad\quad\hspace{0.2cm} +\hspace{0.25cm} \mathbb{E}_{x_t, \mathbf{C}_t, z}[1 - log D(x_t, G(x_t, \mathbf{C}_t, z))] \]  
\[\quad\quad\quad\quad\quad\hspace{0.25cm}+\hspace{0.27cm}\mathbb{E}_{x_t, \mathbf{C}_t, y_t, z}[\left\|y_t - G(x_t, \mathbf{C}_t, z)\right\| _1]\big\}     \quad\quad(2) \]

where $x_t$ are semantic maps, $y_t$ are real sonar images, $z$ is random noise vector, and $ \mathbf{C}_t = [c_{t-1}, \theta_t] $ is a collection of condition variables for the generator. The first two lines of (2) represent the discriminator and generator losses respectively, and the last one is the L1 loss, a regularization term that is meant to discourage blurring in the generator output \cite{pix2pix2016}.

\subsubsection{At test time} only the generator is used and runs identically to the train-time, except $c_t$ now comes from the end of the previous image generated. The model output is therefore dependent on its own previous output and capable of producing consistent and continuous images of any length.

\section{RESULTS AND COMPARISONS} \label{results_section}
\subsection{Experiment 1: Image quality assessment results}
In this experiment we compare MC-pix2pix with real images as well as with the output of the original pix2pix and pix2pix with post-processing, i.e., with blending the border-line between the separate synthetic snippets using sigmoid-function smoothing.
The achieved results are compared both qualitatively and quantitatively as follows:
\subsubsection{Visual examples} of all the methods are provided in Figure~\ref{results_visual}. These are directly comparable as they are generated from the same semantic maps (left), obtained via segmenting the real seabed images (right). Their underlying generative models are trained for the same number of epochs on the same dataset. In order to further eliminate the disadvantage for baseline methods that do not use the yaw variable, no yaw variation was applied in this example (i.e., no turns). This example is primarily illustrating the consistency of the MC-pix2pix output compared to the baselines.
\subsubsection{Visual assessment scores} are obtained from 30 human experts (different levels of experience with sonar data - from introductory course to several years of work with sonar images). Although it is common to use Amazon Mechanical Turk for obtaining such assessment, it is not feasible in our study since real data are both commercially sensitive and too specialized to get a valuable assessment by people previously unexposed to the sonar imagery. Instead we obtain our assessments from the human experts who possess some knowledge of the domain.

The test consists of a number of images generated by MC-pix2pix, pix2pix, sigmoid-blended pix2pix, and corresponding real examples presented in even proportions for a human expert to classify as real or synthetic. The order of images from different models is randomised to avoid putting any of the methods into a disadvantage of being examined last.

Results are presented in Table~\ref{results_human}. Domain experts had 0.52 mean accuracy labelling MC-pix2pix images as real or fake. This is essentially an optimal result because for a two-class problem (real or fake), proximity to 0.5 means experts being as good as random at telling the synthetic data apart from real. Further we present the proportion of synthetic data mislabelled as real (i.e., the success of generator in ``fooling" human experts). For comparison, the proportion of the real data labelled as real is 0.66. The last metric of the visual assessment is the average time taken to make a decision on a sample. Interestingly, participants spend more time on MC-pix2pix images, which suggests these were more challenging to classify. Our method compares favourably to all of the presented baselines for all the presented metrics.

\setlength{\tabcolsep}{0.42em}
\begin{table}

\vspace{0.2cm}
\begin{center}
\ra{1.3}
\begin{tabular}{c c c c c }
\hline
Metrics  &pix2pix &sigma-pix2pix &MC-pix2pix \\
\hline
\hline
\multicolumn{1}{l}{Mean accuracy of labeling} & 0.64 &0.62 & \textbf{0.52} \\
\multicolumn{1}{l}{Synthetic labeled as real} & 0.34  & 0.42 & \textbf{0.54} \\
\multicolumn{1}{l}{Mean time per image (sec)} & 4.85 & 4.86 & \textbf{6.13} \\
\hline
Fresch\'et  Inception  Distance & 0.9257 & 1.0241 & \textbf{0.7834} \\
\hline
\end{tabular}
\end{center}
\caption{\textbf{Image Test Scores:} the average accuracy around 0.5 shows that humans are as good as random at telling MC-pix2pix images apart from real.
MC-pix2pix gets labelled as real more than the competitors and comes the closest to the 0.66 ratio of real images labelled as real.
Image processing times show MC-pix2pix images are the most challenging to inspect.
The lowest FID score confirms the MC-pix2pix is closer to the real image distribution than the competitors.
}
\label{results_human}
\end{table}

\setlength{\tabcolsep}{0.55em}
\begin{table}

\vspace{0.1cm}
\begin{center}
\ra{1.3}
\begin{tabular}{ c c c c c }
\hline
\multicolumn{1}{l}{\textbf{Test: Flat}} & Real only &  + Noise &  + SonarSim &  + MC-pix2pix \\
\hline
\hline
\multicolumn{1}{l}{mAP} & 0.30 & 0.39 & 0.27 & \textbf{0.45} \\
\multicolumn{1}{l}{F1-score} & 0.57 & 0.58 & 0.50 & \textbf{0.60} \\
\hline
\hline
\multicolumn{1}{l}{\textbf{Test: Complex}} & Real only &  + Noise &  + SonarSim &  + MC-pix2pix \\
\hline
\hline
\multicolumn{1}{l}{mAP} & 0.00 & 0.01 & 0.00 & \textbf{0.11} \\
\multicolumn{1}{l}{F1-score} & 0.23 & 0.59 & 0.62 & \textbf{0.68} \\
\hline
\end{tabular}
\end{center}
\caption{\textbf{Bootstrapped ATR performance improvement}: MC-pix2pix improves ATR mAP and F1 score compared to just using real data, as well as beats the baselines for both non-complex flat (top) and complex (bottom) test terrains.}
\label{results_atr}
\end{table}

\subsubsection{Fresch\'et  Inception  Distance (FID)} is also provided at the bottom of the Table~\ref{results_human}. Lower values correspond to the distributions closer to the real one. For instance the FID between two real data sets would be approximately zero, whereas the FID between a constant and $U(0,1)$ is greater than 6. FID is calculated on test images of size 1856$\times$512. This size was chosen arbitrarily - similar results are expected from larger images. FID is sensitive to scaling, so the data for FID assessment has been normalized to the [0, 1] range. 
\subsubsection{Generation speed} the MC-pix2pix is expected to be fairly close to the original pix2pix in speed due to our model being an extension of pix2pix. We used GTX 1080 Ti (12GB RAM) for estimating the MC-pix2pix generation speed. MC-pix2pix is approximately 18 times faster at test time than the real acquisition speed. Marine Sonic acquires 17,100 $\pm$ 10$\%$ pixels per second depending on the settings of the sonar. 



\subsection{Experiment 2: Improving ATR training with MC-pix2pix} \label{atr_section}
\subsubsection{Motivation}
training ATR on simulated data is useful in case of the lack of complexity in training data, or the lack of training data itself, in which case adding more realistic simulated data would be beneficial.
If certain seabed types are underrepresented in the currently available training set, but MC-pix2pix was exposed to these types of terrains before, it can enrich the dataset with additional seabed types.

\begin{figure}
    \centering
    
    \vspace{0.2cm}
    \begin{overpic}[scale=0.95]{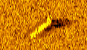}\put(3,40){\textbf{\textcolor{white}{\circled{1}}}}\end{overpic}\hspace{0.06cm}\begin{overpic}[scale=0.685]{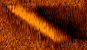}\put(3,40){\textbf{\textcolor{white}{\circled{2}}}}\end{overpic}\hspace{0.06cm}\begin{overpic}[scale=0.685]{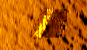}\put(3,40){\textbf{\textcolor{white}{\circled{3}}}}\end{overpic}\hspace{0.06cm}\begin{overpic}[scale=0.685]{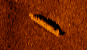}\put(3,40){\textbf{\textcolor{white}{\circled{4}}}}\end{overpic}

    \vspace{0.06cm}
    
    \includegraphics[scale=0.95]{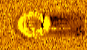}\hspace{0.06cm}\includegraphics[scale=0.685]{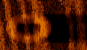}\hspace{0.06cm}\includegraphics[scale=0.685]{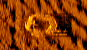}\hspace{0.06cm}\includegraphics[scale=0.685]{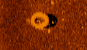}
    
    \caption{\textbf{Examples of target objects (tyres and cylinders) for the ATR training:} 
    1. Random uniform noise background, 2. SonarSim\textsuperscript{2}, 3. MC-pix2pix, and 4. Real data. Objects in pictures 1-3 are synthetic, inserted with the \cite{SSPDcycleGAN}\textsuperscript{3} method. 
    }
    \label{contacts}
\end{figure}

\subsubsection{Experiment goals and the ATR network}
in both of these cases we check the increase in the ATR performance between training on just a small real dataset and enriching it with MC-pix2pix. We assess the performance with mAP and F1-score, as explained in Sec.~\ref{f1}. We are interested in the increase in the ATR performance only - the performance level itself is irrelevant here. The ATR method used in this test is a generic RetinaNet-type network \cite{retina2017}.

\subsubsection{Baselines explained} we train 4 ATR networks on the corresponding datasets: real data only (flat and non-complex), the same real data plus MC-pix2pix images, and baselines - real dataset plus uniform random noise backgrounds, and real dataset plus SonarSim seabeds\footnote{SonarSim - standard vaguely realistic side-scan simulator as used in \cite{SSPDcycleGAN}, capable of generating various seabed textures with limited user control over the type of generated data, but not the exact topography.}. All except the real data are augmented with synthetic targets using the method from \cite{SSPDcycleGAN}\footnote{Due to extremely low amount of the training data for targets we could not generate targets with MC-pix2pix.}, examples of these are provided in Fig.~\ref{contacts}.

\subsubsection{Experiment 2.1: Data shortage} for this experiment MC-pix2pix was trained on the available real training set (flat and non-complex). The MC-pix2pix-generated data were used to train the ATR, which then was tested on another flat and non-complex dataset. Table~\ref{results_atr} (top) shows that MC-pix2pix provides significant improvements in MAP, compared to just real data and other baselines, and the best F1.

\subsubsection{Experiment 2.2: Lack of complexity} in this case MC-pix2pix was pre-trained with slightly more complex ripply seabeds, emulating previous exposure to the complex data. It then generated more of the complex seabeds, that were used to train the ATR alongside the flat and non-complex real data. When testing this ATR on complex real terrains (results presented at the bottom of Table~\ref{results_atr}), both F1-score and mAP drastically improve with the MC-pix2pix data bootstrapping, compared to just real data training and baselines.

This confirms that MC-pix2pix could be deployed as a highly efficient bootstrapping technique for improving ATR performance in cases of low real data availability or low real data diversity, that are common in the real life applications.

\vspace{-4px}
\subsection{Addressing Potential Concerns:}
\subsubsection{Advantages compared to generating the seabed piece-wise and then stitching it together with post-processing} Although standard smoothing techniques like sigmoid-smoothing interpolate well between the colours, they do not conduct the texture integration between the images as is evident from the Fig.~\ref{results_visual} middle section (at border-line).  
        
        
        

    \begin{figure}
    
        \vspace{0.2cm}
        \flushleft
        {$\Hquad$\textcolor{gray}{$\quad$Starboard$\quad\quad\quad\quad\quad\quad\quad\quad$Port}}
        
        \centering
        \vspace{0.05cm}
        \rotatebox{90}{$\Hquad$\textcolor{gray}{maps \& yaw}}\hspace{0.05cm}\includegraphics[scale=0.315]{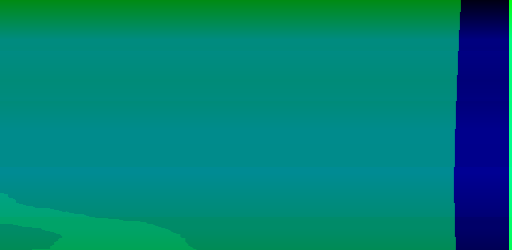}\includegraphics[scale=0.315]{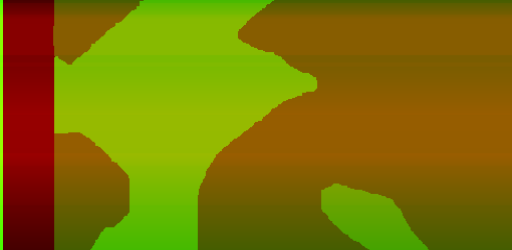}
        
        \vspace{0.05cm}
        \rotatebox{90}{$\Hquad$\textcolor{gray}{pix2pix}}\hspace{0.05cm}\includegraphics[scale=0.315]{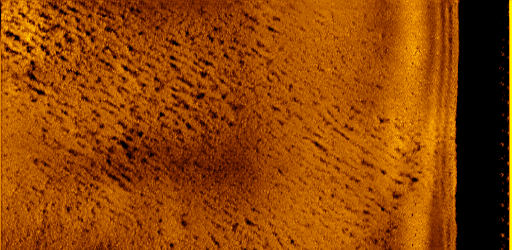}\includegraphics[scale=0.315]{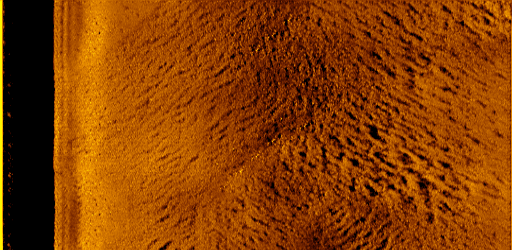}
        
        \vspace{0.05cm}
        \rotatebox{90}{$\Hquad$\textcolor{gray}{MC-pix2pix}}\hspace{0.05cm}\includegraphics[scale=0.315]{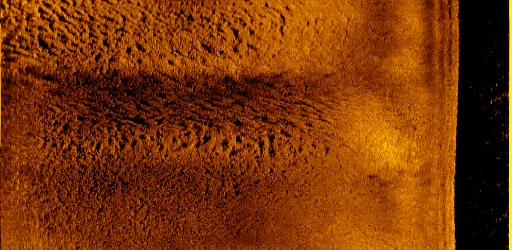}\includegraphics[scale=0.315]{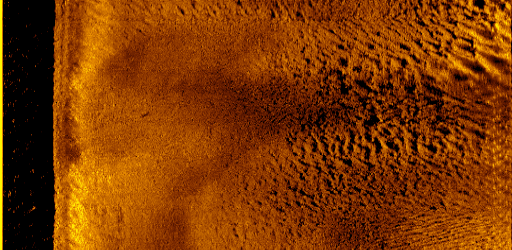}
        
        \vspace{0.05cm}
        \rotatebox{90}{$\Hquad$\textcolor{gray}{Real}}\hspace{0.085cm}\includegraphics[scale=0.315]{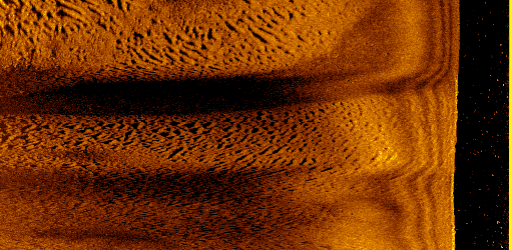}\begin{overpic}[scale=0.315]{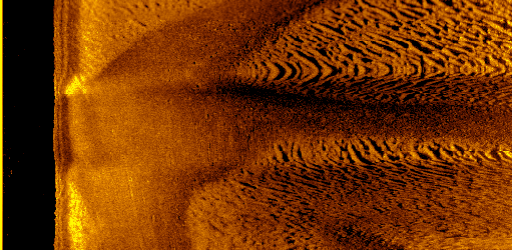}\put(60,8){\includegraphics[scale=0.1]{scheme-short.png}}\put(66, 3){\footnotesize{\textcolor{peach}{10 meters}}}\end{overpic}

        \caption{\textbf{Reproducing turns with yaw-conditioning (top-to-bottom):} 0. semantic map overlayed with yaw. 1. pix2pix fails to capture a distortion caused by vehicle turning as semantic map provides no indication of turns other than perhaps indirectly through the topography. 2. MC-pix2pix benefits from a inbuilt yaw-based condition, getting close to real turn patterns and distinguishing between the inside (left) and the outside (right) turns. 3. Real example.}
        \label{turns}
        \vspace{-3px}
    \end{figure}

\subsubsection{Quality Decay over the generation time} self-conditional model at test time suggests that the quality drop could accumulate over time. However, the nature of GAN architecture prevents this - trained GAN always samples the training data distribution, regardless of the condition. This has been verified via generating a standard test mission - average duration of 2 hours, 300,000 pixels along the track.
\subsubsection{Handling vehicle turns}  we condition on the yaw only (because roll and pitch are not available in our data). It does not seem to be quite enough information to make results fully realistic, however the model not only acknowledges the concept of a turn distortion but also distinguishes between the inside and the outside turns, in some cases producing very realistic results (especially successful for the outside turns simulation). A visual example of what real vs. generated turns look like is presented in Fig.~\ref{turns}. 
\subsubsection{Handling multiple viewpoints} 
\begin{figure}
        \vspace{0.2cm}
        \flushleft
        {\textcolor{gray}{$\Hquad$Layout$\quad\quad\quad$View 1$\quad\quad\quad\Hquad$View 2$\quad\quad\quad\quad$Overlay}}

        \centering
        \includegraphics[scale=0.123]{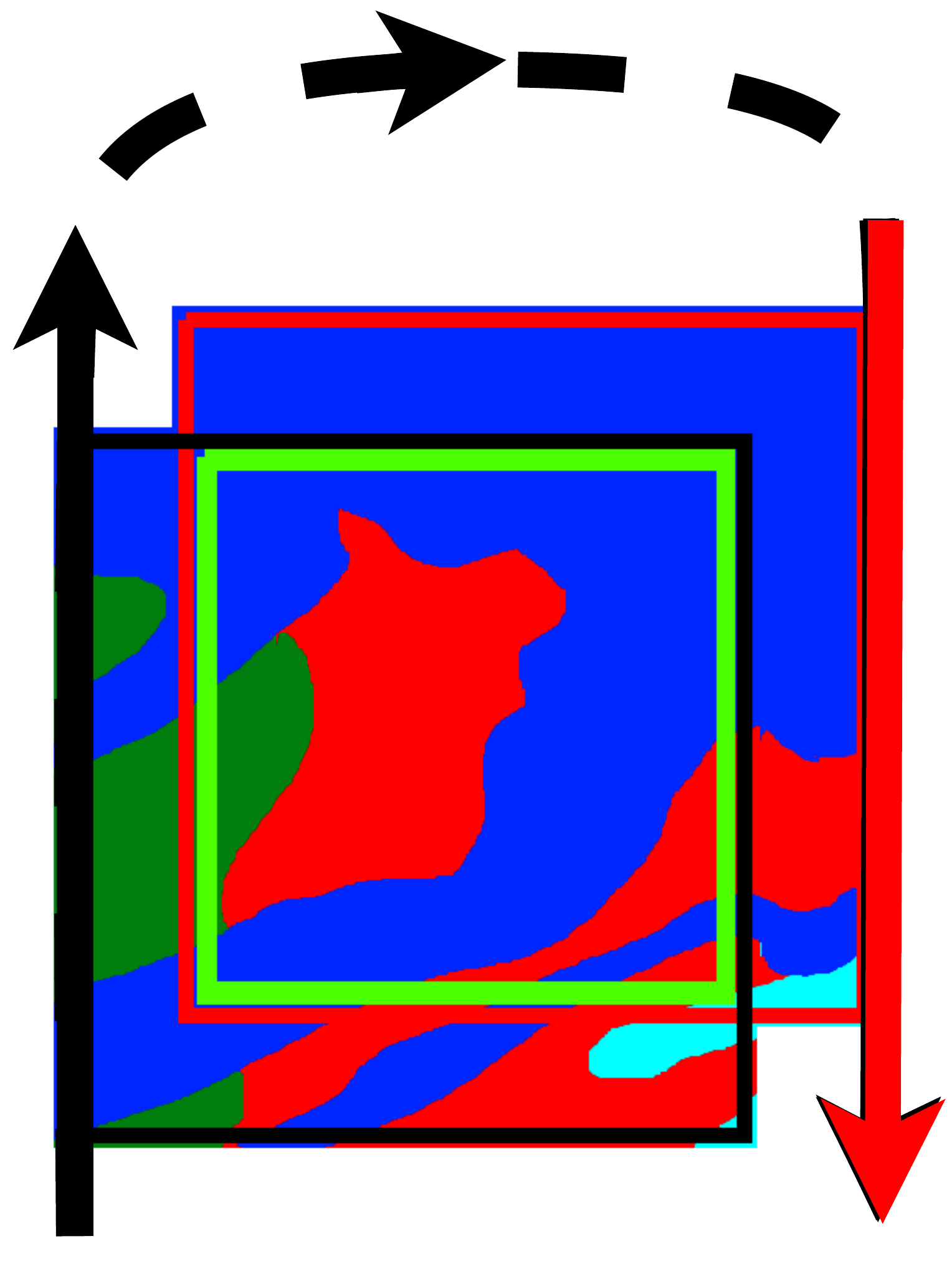}\hspace{0.06cm}\includegraphics[scale=0.175]{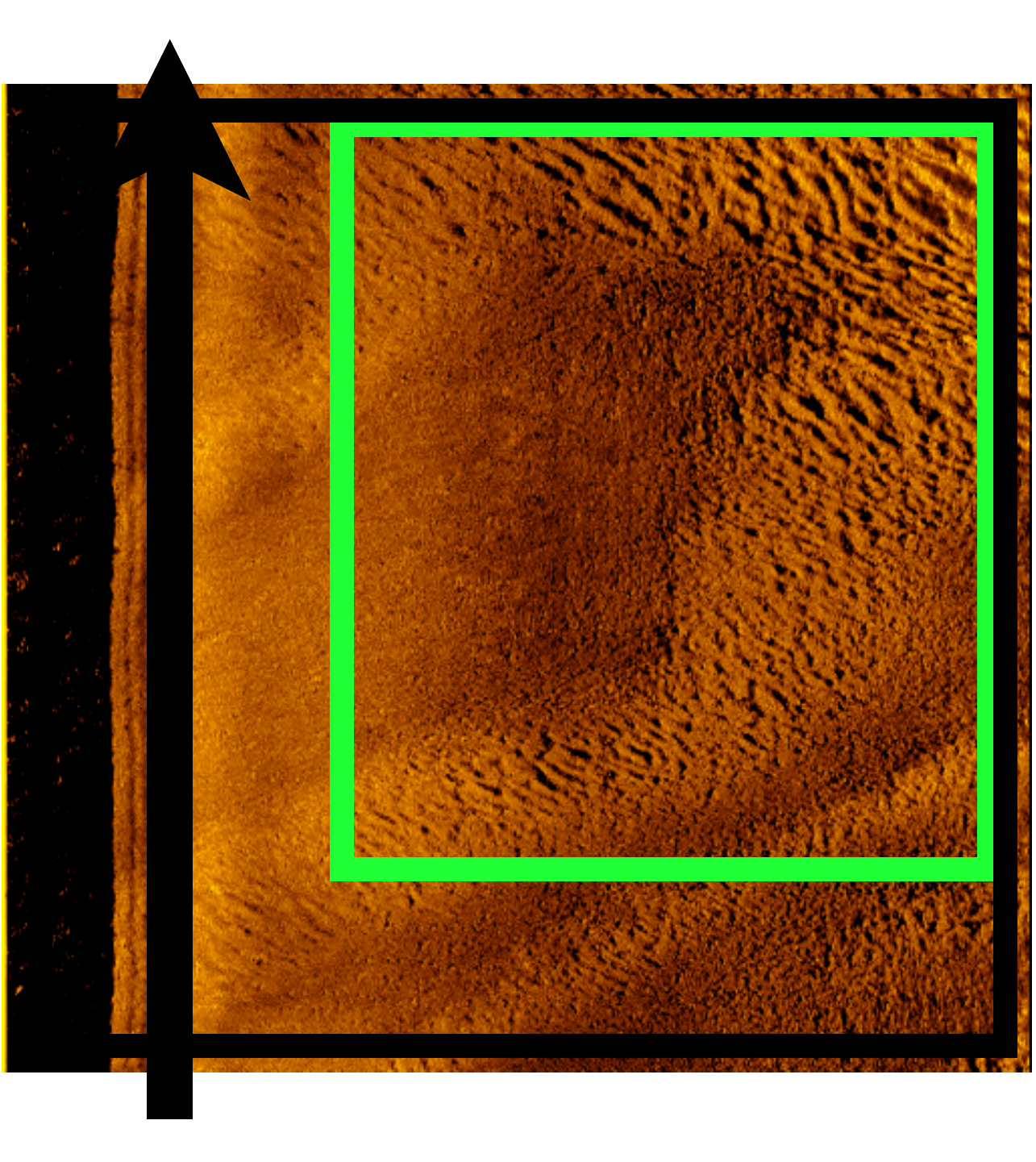}\hspace{0.06cm}\includegraphics[scale=0.175]{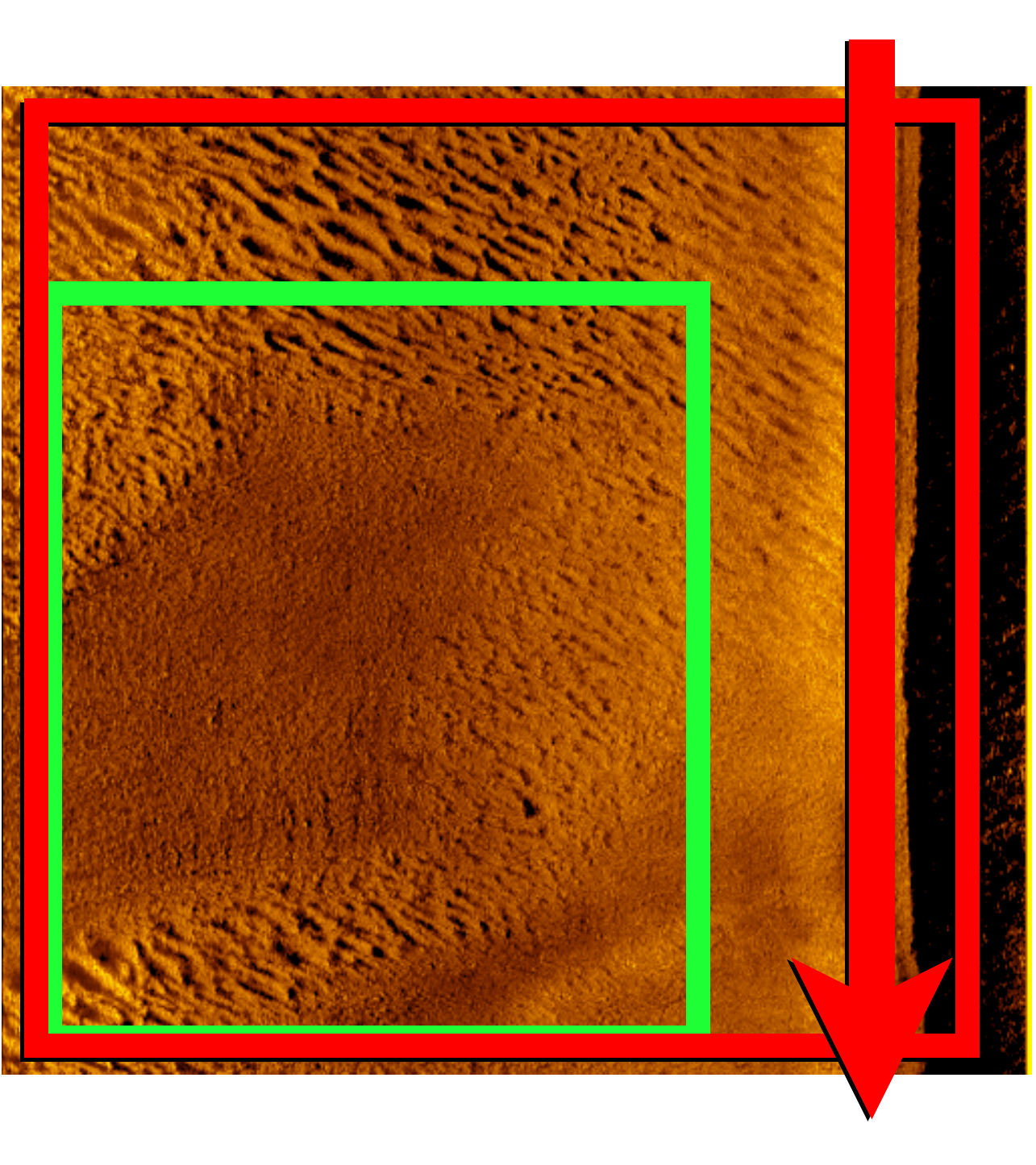}\hspace{0.06cm}\includegraphics[scale=0.21]{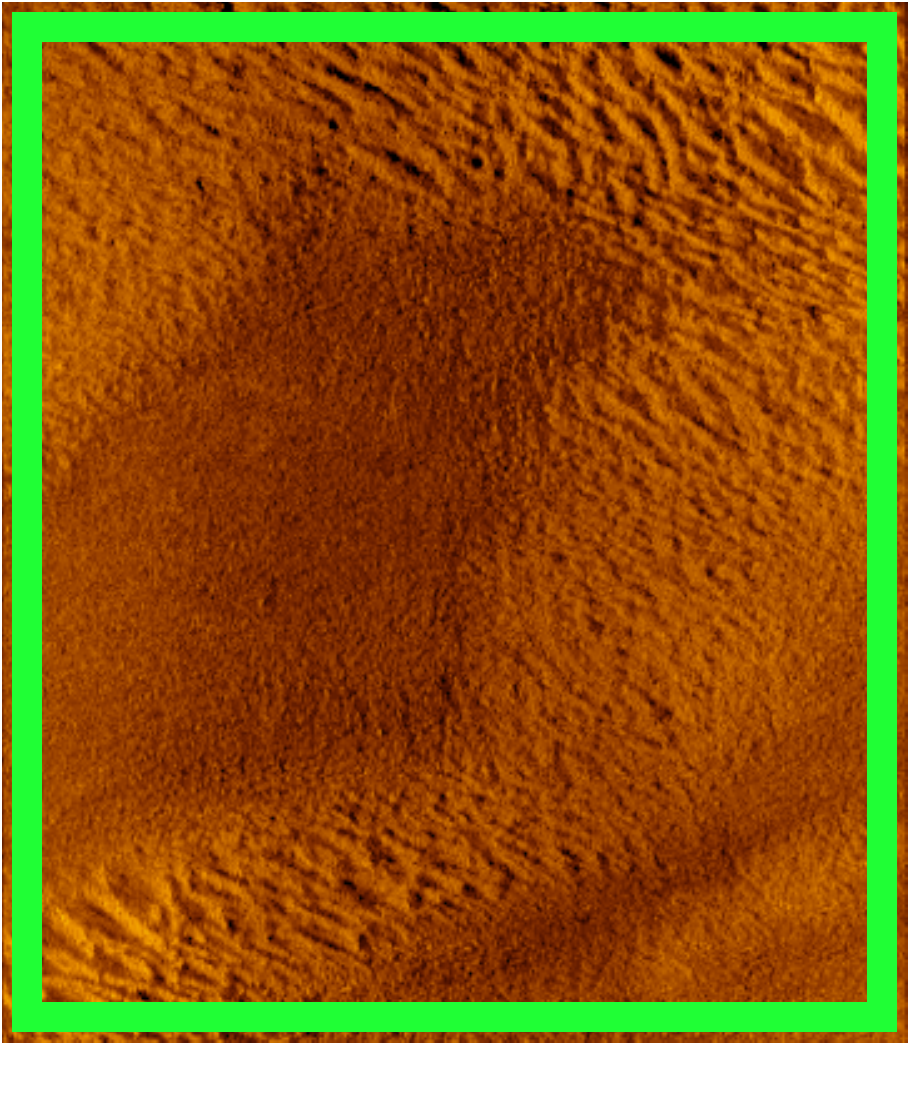}
        
        
        \vspace{-3px}
        \caption{\textbf{Handling different view-points:} MC-pix2pix is topographically consistent with respect to the types of terrains. This example shows the same region, generated as perceived from two different viewpoints with some displacement. Generated images show a clean overlap without contradictions.
        }
        \vspace{-3px}
        \label{viewpoints}
    \end{figure}

typically a vehicle observes the same area at least twice. The MC-pix2pix accounts for the topographical coherence with respect to the terrain types. The model naturally produces a similar image for the different viewpoints. Example in Figure~\ref{viewpoints} shows two views of the same area (e.g. the image synthesized for the same map approached from different sides) and their overlay.
\subsubsection{Unconditional generation} this work focuses on producing the missions with user-controlled topography, however if one needs to avoid specifying it (e.g. when producing the training data for ATR) the original GAN \cite{vanillaGAN} and modifications, such as DCGANs \cite{DCGANs}, can be employed to generate some semantic maps for the input into MC-pix2pix.
\subsubsection{Other potential applications} MC-pix2pix can be potentially used in any setting where simulation of large-scale continuous data is required. Useful for bootstrapping learning algorithms, environment feature detection and recognition, or even for side-scrolling games background generation.

\section{CONCLUSIONS AND FUTURE WORK}
This work proposes a method for generating realistic synthetic sonar sensory data for full-length underwater missions with a direct control over the topography. To our knowledge this is the first published work that addresses generation of side-scans for entire missions with generative adversarial networks. Examples of visual results were provided along with the quantitative assessment of the model. These include FID scores, generation speed, ATR performance improvement results when bootstrapped with the MC-pix2pix generated data, and visual assessment scores confirming MC-pix2pix synthetic data looks very realistic to humans.

In future work we will investigate the use of roll and pitch data for improving quality of the simulation for the vehicle turns, subject to the availability of the suitable training data.

The main extension to this work is generation of the data for higher fidelity sonars, such as EdgeTech (an order of magnitude higher resolution compared to the Marine Sonic data presented in this work), or SAS sonars (two orders of magnitude higher in resolution compared to Marine Sonic). Despite being very challenging this problem can be addressed with some limited extensions to the current MC-pix2pix algorithm and is currently a work in progress.

\addtolength{\textheight}{-13cm}   



\section*{ACKNOWLEDGMENT}

We thank Stephanos Loizou and Peter Scanlon for their help with the ATR experiments, and Roshenac Mitchell, Joshua Smith, and Henry Gouk - for the technical support; as well as all the participants of visual assessment experiments.

\bibliographystyle{IEEEtran}
\bibliography{IEEEexample}

\end{document}